# Machine learning with incomplete datasets using multi-objective optimization models

Hadi A. Khorshidi
School of Computing and Information Systems
The University of Melbourne
Melbourne, Australia
hadi.khorshidi@unimelb.edu.au

Michael Kirley
School of Computing and Information Systems
The University of Melbourne
Melbourne, Australia
mkirley@unimelb.edu.au

Uwe Aickelin
School of Computing and Information Systems
The University of Melbourne
Melbourne, Australia
uwe.aickelin@unimelb.edu.au

*Abstract*—Machine learning techniques have been developed to learn from complete data. When missing values exist in a dataset, the incomplete data should be preprocessed separately by removing data points with missing values or imputation. In this paper, we propose an online approach to handle missing values while a classification model is learnt. To reach this goal, we develop a multi-objective optimization model with two objective functions for imputation and model selection. We also propose three formulations for imputation objective function. We use an evolutionary algorithm based on NSGA II to find the optimal solutions as the Pareto solutions. We investigate the reliability and robustness of the proposed model using experiments by defining several scenarios in dealing with missing values and classification. We also describe how the proposed model can contribute to medical informatics. We compare the performance of three different formulations via experimental results. The proposed model results get validated by comparing with a comparable literature.

## I. INTRODUCTION

Most of machine learning techniques have been developed based on the assumption of having complete data [1]. While existence of missing values is a reality in data analysis especially in medical informatics. The datasets can be incomplete due to failure in measurement equipment, reluctance or lack of knowledge in providing response in surveys, incomplete investigation, data collection errors, etc. There are three major types of missing data as missing at completely random (MCAR), missing at random (MAR) and missing not at random (MNAR) [2]. The simplest way to handle missing values and create a complete data is to discard records with missing values. However, the records with missing values hold valuable information, and removing them leads to unreliable analysis results [3]. In addition, in many cases e.g. cancer research, collecting data is time consuming and costly and every record is highly valuable. Therefore, imputation methods are introduced to estimate the actual values of the missing values. Imputing the missing values plays an important role to develop unbiased machine learning models [4].

Many imputation methods have been developed to convert incomplete datasets into reliable complete data. However, applying imputation methods is an offline process for machine learning. As shown in Fig. 1, several steps should be taken for prediction in a case the train and test datasets have missing values. These steps are I. Imputation for train data, II. Training the model, III. Imputation for test data and IV. Prediction using trained model. In addition, steps I and II need efforts on parameter tuning and model selection. Moreover, we cannot be certain which imputation method is the best, how accurate the imputation is and how the imputation contributes toward prediction.

There are few studies that have attempted to develop machine learning models for incomplete datasets without imputation. In these studies ([5],[6]), researchers estimate the distance between incomplete feature vectors for distance-based supervised learning. In our study, we propose an online approach for machine learning of incomplete data using a multi-objective optimization. In other words, the proposed machine learning method perform imputation, training and prediction (steps in Fig. 1) simultaneously. In addition, this method completes parameter tuning and model selection at the same time to achieve optimal imputation and prediction.

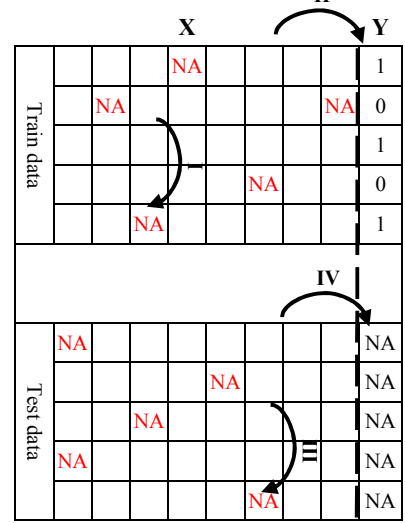

Fig. 1. Conventional process of imputation and learning

Multi-objective optimization modeling has been used in machine learning for semi-supervised clustering [7],[8], feature selection [9], model selection [10], model pruning [11], train subset selection (TSS) [12]. However, it has not been used for imputation and prediction simultaneously, to the best of our

knowledge. The research questions (RQs) in our study are as below:

RQ1. How to formulate the objective functions which can properly represent imputation and prediction?

RQ2. How to find the optimal solutions?

RQ3. How robust and reliable are the optimal solutions?

We address these research questions in sections III and IV. In our study, we choose classification as the predication task. We also use support vector machine (SVM) as the classification technique. It is because SVM is a powerful machine learning technique but its performance is dependent on the combination of selected parameters and SVM has been a subject of optimization modeling for model selection [13]. In addition, we choose fuzzy clustering for imputation of missing values. The fuzzy clustering is selected because it is a validated imputation method [14, 15] and it provides smarter way to formulate solutions for finding optimal solutions in comparison with using all missing values as decision variables [16].

## II. PRELIMINARIES

### A. Support Vector Machine (SVM)

SVM is a supervised statistical learning method which is introduced by Vapnik [17] and can be used for both classification and regression tasks. In support vector classifier, the main objective of SVM is to find the optimal separating hyperplane by maximizing the distance of the hyperplane from support vectors as (1). Support vectors are the data points from both classes which have the minimum distance from hyperplane [18]. This procedure makes the classifier robust as it is just sensitive to change in support vectors not all data points.

$$Max\ M \quad (1)$$

Subject to

$$y_i(w.\emptyset(x_i) + b) \geq M(1 - \epsilon_i)$$

$$\epsilon_i \geq 0, \sum_{i=1}^{n} \epsilon_i \leq C$$

where $x_i$ is data point $i$, $y_i$ is its label, $n$ is the number of data points and $\emptyset(.)$ is the function to map the original space into a high-dimensional space where constructing a separating hyperplane is possible [19]. Also, $C$ gives flexibility to have wrong classified data points by defining a margin for the hyperplane, $M$ is the width of the margin and $\epsilon_i$ is the slack variable that allows data point $i$ be on the wrong side of the margin or the hyperplane.

To avoid the complexity of computing in high-dimensional space, kernel functions are introduced. Kernel functions directly apply in the original space [11]. The most popular kernel functions are listed in TABLE I.

To reach the best SVM model, the optimal combination of parameter $C$, the kernel function ($Kr$) and its parameters $\gamma, r$ and/or $d$.

TABLE I. KERNEL FUNCTIONS

| Linear Kernel | $k(x_i, x_j) = x_i . x_j$ |
|---|---|
| Radial Kernel | $k(x_i, x_j) = \exp(-\gamma \|x_i - x_j\|^2)$ |
| Polynomial Kernel | $k(x_i, x_j) = (\gamma(x_i . x_j) + r)^d$ |
| Sigmoid Kernel | $k(x_i, x_j) = tanh(\gamma(x_i^T . x_j) + r)$ |

### B. Fuzzy clustering

Fuzzy clustering is a soft allocation clustering method which is introduced by Bezdek [20]. In soft allocation, a data point does not belong to just one cluster. Similarly in fuzzy clustering, a data point allocates to all clusters with degrees of membership which denote how much the data point belongs each cluster [14]. Fuzzy clustering is to minimize the objective function formulated in (2).

$$\sum_{i=1}^{n} \sum_{k=1}^{c} u_{ik}^v \|x_i - c_k\|^2 \quad (2)$$

where $c$ is the number of clusters, $c_k$ denotes center of cluster $k$, $u_{ik}$ is the membership degree for data point $i$ in cluster $c$, and $v$ is the fuzziness parameter. Once $c$, $m$ and cluster centers are determined, the membership degrees are calculated as (3) [21]. These parameters have impact on the performance of fuzzy clustering and should be tuned properly [22].

$$u_{ik} = \begin{cases} \left(\sum_{j=1}^{c} \left(\frac{\|x_i - c_k\|^2}{\|x_i - c_l\|^2}\right)^{\frac{1}{v-1}}\right)^{-1} & if\ \|x_i - c_l\|^2 > 0 \\ 1 & if\ \|x_i - c_k\|^2 = 0 \\ 0 & \exists l \neq k\ if\ \|x_i - c_l\|^2 = 0 \end{cases} \quad (3)$$

## III. PROPOSED MULTI-OBJCTIVE MODEL

In this section, we describe all elements of the proposed multi-objective optimization model including objective functions, the procedures of imputation, model selection and solving algorithm which is based on Non-dominated Sorting Genetic Algorithm (NSGA) II. The proposed model is to skip steps mentioned in Fig. 1 and perform imputation and model selection at the same time. As shown in Fig. 2, features ($X$) and labels ($Y$) of the train dataset are used for model selection, and in parallel features of train and test datasets are used for imputation.

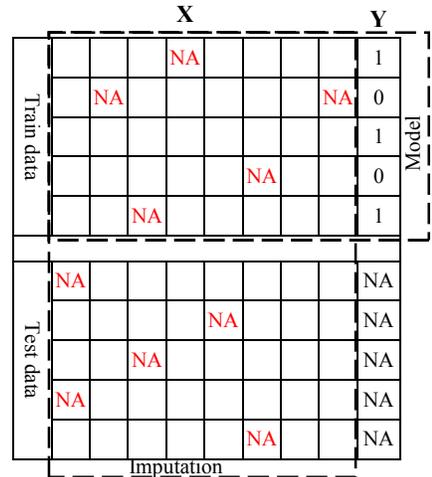

Fig. 2. Proposed imputation and learning

## A. Objective functions for imputation

The first challenge to develop the multi-objective optimization model is to formulate an objective function for imputation task. The performance of the imputation normally is evaluated using the difference between the actual and imputed values by root mean square error (RMSE) and mean absolute error (MAE). However, the actual values of the missing data are not available. So, we need to formulate an objective function that represents the imputation error (addressing RQ1). In this paper, we propose three objective functions for imputation, and examine their performance by experiments.

### 1) Cluster validity function

Cluster validity indices are used to evaluate the performance of a clustering task and tune the parameters of clustering methods [23]. As the imputation method (in our study) is based on fuzzy clustering, cluster analysis indices are proposed as an objective function to examine the imputation through the optimization process. In the literature, cluster validity indices have been used to determine the number of clusters for cluster-based imputation methods [14]. Here, we formulate the first objective function as average silhouette width (ASW) in (4). ASW is a cluster validity index which measures how much data points within each cluster are similar and are dissimilar from data points in other clusters.

$$ASW = \frac{1}{n}\sum_{i=1}^{n} s(i,k) \quad (4)$$

where $s(i,k) = \frac{b(i,k)-a(i,k)}{max\{a(i,k),\ b(i,k)\}}$, $a(i,k)$ is the average dissimilarity between data point $i$ and other data points in the same cluster, and $b(i,k)$ is the minimum average dissimilarity between data point $i$ and other data points in clusters to which data point $i$ does not belong. ASW is the higher the better index [24].

### 2) Correlation function

Canonical correlation analysis (CCA) is a powerful multivariate data analysis technique to find the correlation between two datasets. It has been used in transfer learning by maximizing the correlation of transferred source and target datasets using as (5).

$$corr(X_s, X_t) = \frac{w_s^T cov(X_s,X_t) w_t}{\sqrt{(w_s^T var(X_s) w_s)(w_t^T var(X_t) w_t)}} \quad (5)$$

where $X_s$ and $X_t$ are source and target datasets respectively, $w_s$ and $w_t$ are the weight vectors that should be optimally determined to maximize the correlation of transformed source and target datasets, and $T$ denotes transverse [25].

We adopt this concept to formulate the second objective function for imputation. We expect the correlation of train dataset ($X_{tr}$) and test dataset ($X_{te}$) should be maximized if the imputation is done accurately. Also, as these datasets are from the same space, we do not need to determine the weight vectors. So, we replace them with vectors of 1 and reformulated (5) as (6) for imputation.

$$corr(X_{tr}, X_{te}) = \frac{1_{tr}^T cov(X_{tr},X_{te}) 1_{te}}{\sqrt{(1_{tr}^T var(X_{tr}) 1_{tr})(1_{te}^T var(X_{te}) 1_{te})}} \quad (6)$$

### 3) Variance ratio function

The third objective function is formulated based on the assumption that the variance of train and test datasets should be the same as they have similar characteristics. So, we formulate variance ratio (VR) as (7). If the missing values are imputed properly, we expect that VR gets higher values and becomes closer to one.

$$VR(X_{tr}, X_{te}) = \frac{min\left((1_{tr}^T var(X_{tr}) 1_{tr}),(1_{te}^T var(X_{te}) 1_{te})\right)}{max\left((1_{tr}^T var(X_{tr}) 1_{tr}),(1_{te}^T var(X_{te}) 1_{te})\right)} \quad (7)$$

## B. Objective function for model selection

Formulation of an objective function for model selection is a simpler task. In machine learning, classification models are created using train datasets to be used for prediction in test datasets. To assess the prediction performance, several techniques have been introduced. Cross-validation is an experimental design technique to estimate the prediction error of test dataset based on train dataset [18]. In this paper, we use 10-fold cross-validation to calculate the classification error of train dataset using SVM as an objective value for model selection.

As a result, we have a bi-objective optimization model to maximize the imputation and minimize the model selection objective functions respectively.

## C. Imputation and model selection procedure

### 1) Decision variables and initialization

To reach the goal of the proposed multi-objective optimization model, we need to find the optimal combinations of the parameters for imputation and classification. As the imputation is based on fuzzy clustering, the parameters are the number of clusters, center points and fuzziness. For classification, the best SVM model is selected by tuning the flexibility parameter, finding the kernel function and its parameters. Therefore, we will have a decision variable vector, which is called chromosome in genetic algorithm (GA), as shown in Fig. 3. Each decision variable in the chromosome is called gene. Note that $m$ is the number of features in each dataset.

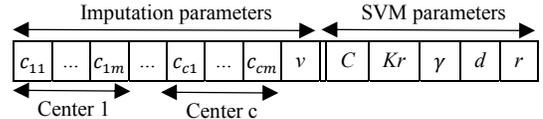

Fig. 3. Chromosome design

To initialize the procedure, a population with $P$ chromosomes are generated. The length of each chromosome depends on the number of clusters. In this study, we let the number of clusters be variable, we just set the maximum number of clusters ($K$). So, each population includes chromosomes with different lengths which denote clustering with different numbers of clusters. The number of clusters will be optimized through the procedure.

### 2) Imputation

Once cluster's centers and fuzziness parameters are generated (or updated), the membership degree of each data point for each cluster can be calculated using (3). For the first population, the missing values are initially imputed using mean imputation before membership degree calculation. Once the membership degrees are obtained, the imputed values of the missing values are updated using (8).

$$ImpV_{ij} = \sum_{k=1}^{c} u_{ik} \times c_{kj} \qquad (8)$$

where *(i, j)* denotes that feature *j* in data point *i* is missing.

*3) Objective value calculation*

After imputation, we can calculate the objective values for each chromosome. For imputation objective functions, if the objective function is ASW, the allocated cluster is determined using membership degree so that the bigger membership degree shows to which cluster the data point is belonged. Then, ASW value is calculated using (4) for the chromosome. If the objective function is correlation or VR, the imputed train and imputed test datasets are used to calculate objective values using (6) or (7) respectively.

For the objective value of the selection model, the imputed train dataset is used to develop an SVM model using parameters indicated in the chromosome. The kernel parameters are considered if they are applicable for the kernel function. Then, the cross-validated classification error is calculated as described in section III.B.

*D. Solving algorithm*

In this section, we describe what is the algorithm to find optimal solutions for the optimization model (addressing RQ2).

*1) Pareto fronts*

As the model is a multi-objective optimization one, we need to determine Pareto optimal solutions in each population. The Pareto optimal solutions are the chromosomes that are not dominated by any other chromosome in terms of objective values [26]. The Pareto solutions create a boundary in each population which is called Pareto front. We solve the optimization model using NSGA II which is one of the most efficient multi-objective evolutionary algorithms. In NSGA II, the population is sorted based on successive Pareto fronts for comparison, selection, crossover and mutation operations [27].

In this paper, for each population, we firstly find the Pareto solutions which we call them the first Pareto front. Then, for the rest of chromosomes, we find their Pareto front. This procedure continues until all chromosomes are in Pareto fronts as shown in Fig. 4. The chromosomes in the first Pareto front are moved to the next population with no change. The chromosomes in the last Pareto fronts are replaced by offspring created by crossover operator.

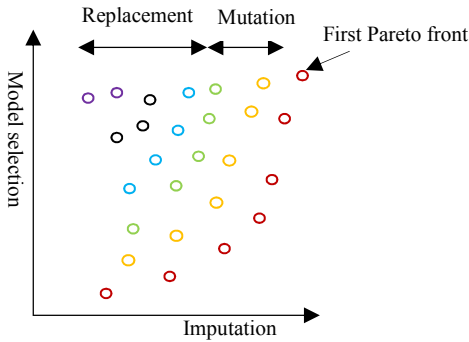

Fig. 4. Pareto fronts

*2) Crossover operator*

For crossover, we select *Cr* number of chromosomes randomly from all chromosomes in the population. Among them, we select two chromosomes which have the best value of each objective function. Also, we select two chromosomes which have the worst value of each objective function. These four chromosomes are parents to generate offspring. We select the chromosomes with the worst values to give all chromosomes a chance to be part of generating the next population. In anti-learning [28] and opposition-based learning [29] concept, it is argued that the instances with the worst values can be used to learn toward optimal solutions.

Once the parents are selected, the parents with best values crossover to generate an offspring. Also, the parents with best values do crossover with their opposite worst parents. For example, the parent with the best value of imputation objective function does crossover with the parent with the worst value of the objective value of the model selection. As a result, three offspring would be generated from each crossover operation.

When two parents do crossover, as the length of chromosomes are different, it firstly should be decided what the length of the offspring would be. We randomly select one of the parents to be the same size of the offspring. Then, a random vector is generated which has binary values. The length of the random vector is equal with minimum length of parent. Fig. 5 shows an example of crossover of two parents which parent 1 and 2 have three and two center points respectively.

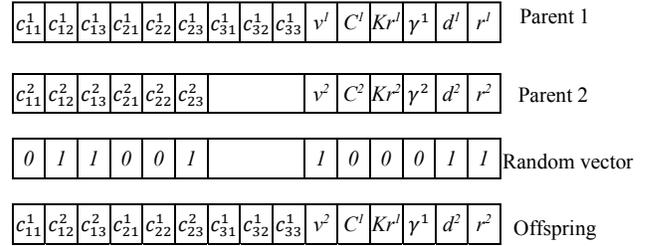

Fig. 5. Crossover operation

*3) Mutation operator*

The chromosomes, which are not in the first Pareto front and are not replaced by offspring, are moved to the next population using mutation. The number of these chromosomes is defined approximately half of the chromosomes where are not in the first Pareto front.

To perform the mutation, firstly, we randomly select how many and which genes would be mutated. Then, a new random value is generated for each gene within the accepted range. Fig. 6 provides an illustrative example on mutation operation.

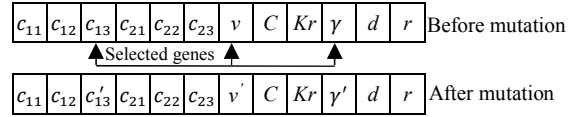

Fig. 6. Mutation operation

*4) Stopping criteria*

The generation of new population stops once one of the below conditions are met.

- The number of generations reaches the maximum number of generations (*tau-max*).
- The difference of the average values of the imputation objective function for all chromosomes in a population in two successive generations is less than a predefined threshold (*thresh*).
- The difference of the average values of the model selection objective function for all chromosomes in a population in two successive generations is less than a predefined threshold (*thresh*).
- All the chromosomes of a population are in the first Pareto front.

## IV. EXPERIMENTS

In this section, we implement the proposed multi-objective optimization model on real datasets and design several experiments to compare different formulations of the model (addressing RQ1) and examine how well the model performs (addressing RQ3).

We use three datasets listed in TABLE II. These datasets are selected because they have different range of class numbers and number of features. They cover numerical and categorical features.

TABLE II. DATASETS

| Data set | Data points | Features | | Class |
|---|---|---|---|---|
| | | Numerical | Categorical | |
| iris | 150 | 4 | - | 3 |
| zoo | 101 | 1 | 15 | 7 |
| sonar | 208 | 60 | - | 2 |

### A. Missing value generation for experiments

The performance of an imputation method and consequently a classification model is impacted not just by the ratio of missing values but also by their patterns and types [30]. We generate different ratios, patterns and types of missing values for our experiments to compare the performance of the proposed model. We set several missing value ratios as 1%, 5%, 10%, 25% and 50%. In addition, we study three different missing patterns as simple, medium and complex. In a simple pattern, each data point at most has only one missing value. A pattern is medium if a data point has missing values in 2 to 50% of its features. The number of missing values for a data point in a complex pattern is between 50% and 80% of the number of features. We consider two types of missing value as overall and uniformly distributed (UD). In the overall model, missing values are randomly distributed across features. In the UD model, all features have an equal chance of having missing values [31]. So, we consider 30 different combinations for missing value generation in our experiments.

### B. Learning with missing values

We consider two situations that missing values are involved in machine learning. In the first situation, missing values present in both train and test datasets (as shown in Figs 1 and 2). This situation happens when collecting complete data is costly and time-consuming. One medical example of this situation is the research on relationship of breast cancer treatment and fertility [32].

In the second situation, missing values only present in test dataset. In this situation, there are sufficient complete labelled data which are reliable to train machine learning models. However, the data for prediction is not free from missing values. A medical example for this situation is intensive care unit (ICU) stay records in MIMIC-II database [33]. We investigate all combinations of missing values in both situations.

### C. Implementation settings and outputs

On top of 3 datasets, 30 combinations of missing value generation and 2 situations, we implement the multi-objective model using 3 different formulations of imputation objective functions as well. As a result, we come up with 540 different implementations.

For model implementation, we set the number of chromosomes in each population as 54 (*P=54*), the maximum number of clusters as 10 (*K=10*), the number of chromosomes selected for crossover as 8 (*Cr=8*). Also, the fuzziness parameter ($v$) is between 1.5 to 5, the flexibility parameter ($C$) is in the range of 0.01 – 100, kernel functions are as listed in TABLE I. Kernel parameters as $\gamma$, $r$ and $d$ are [0.005, 5], [0, 20] and [2, 5] respectively. For stopping criteria, the maximum number of generations is set as 100 (*tau-max=100*) and the threshold is defined as 0.0005 (*thresh=0.0005*).

For each implementation, we record the solution time (in seconds), the number of chromosomes in the first Pareto front in the last population and their objective in average. The summary of the outputs for both situations are presented in TABLE III.

TABLE III. OUTPUT SUMMARY

| Categories | Sub-categories | Test missing | | | Train & test missing | | |
|---|---|---|---|---|---|---|---|
| | | Time (s) | Pareto front | Classification error | Time (s) | Pareto front | Classification error |
| Dataset | Iris | 34 | 52 | 0.03 | 252 | 40 | 0.05 |
| | Zoo | 26 | 53 | 0.04 | 112 | 50 | 0.06 |
| | Sonar | 247 | 52 | 0.17 | 999 | 33 | 0.18 |
| Objective function | ASW | 111 | 52 | 0.08 | 501 | 40 | 0.1 |
| | Correlation | 100 | 52 | 0.08 | 431 | 43 | 0.1 |
| | VR | 95 | 52 | 0.08 | 430 | 40 | 0.1 |
| Ratio | 1% | 93 | 52 | 0.08 | 200 | 51 | 0.08 |
| | 5% | 105 | 52 | 0.08 | 369 | 48 | 0.09 |
| | 10% | 101 | 52 | 0.08 | 470 | 43 | 0.08 |
| | 25% | 101 | 53 | 0.08 | 623 | 31 | 0.1 |
| | 50% | 110 | 52 | 0.08 | 623 | 31 | 0.11 |
| Pattern | Simple | 90 | 52 | 0.08 | 343 | 47 | 0.08 |
| | Medium | 108 | 52 | 0.08 | 508 | 38 | 0.1 |
| | Complex | 108 | 52 | 0.08 | 516 | 36 | 0.12 |
| Type | Overall | 104 | 52 | 0.08 | 479 | 41 | 0.09 |
| | UD | 100 | 52 | 0.08 | 433 | 40 | 0.1 |

Due to space limitation, we present the outputs in an aggregated manner. It means the figures are the average of output values for each sub-category. We also examine the impact of different datasets, imputation objective functions and missing combinations on the outputs statistically. We use Kruskal-Wallis which a non-parametric statistical test.

The statistical test's results are consistent with output summaries in TABLE III. This table provides a comparative insight over the performance of the multi-objected model across different sub-categories within each category. The implementation time, pareto front number and cross-validated classification error in average remain stable (in confidence level of 95%) over changes unless change in data. It seems the model get slower as the number of records increases. The model performs robust in classification in case of different missing combinations. In comparison between two situations, when there are missing values in both train and test data, time of implementation is more, the number of chromosomes in the first Pareto front is less and cross-validated classification error is slightly higher.

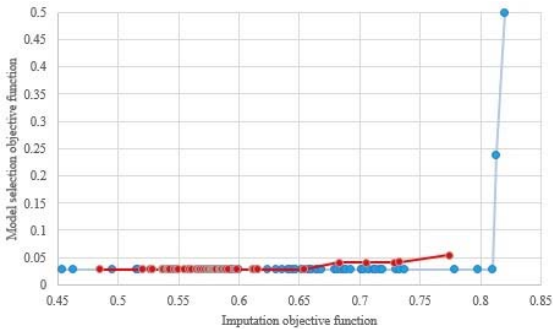

Fig. 7. Examples of Pareto fronts

We also measure the hypervolume of the first Pareto front for each implementation. Fig. 7 shows two first Pareto fronts of the implementations on iris dataset using ASW formulation. The blue points denote the Pareto front with highest hypervolume value, while the red points show the Pareto front with lowest hypervolume value.

*D. Model reliability*

The main purpose of model is to impute missing values accurately in a way that provide reliable prediction. To measure this, for each implementation, we compute the average of test classification error for the first Pareto front in the last population. This classification error is computed under two circumstances 1) when there is no missing value i.e. complete test data are used in the selected model, 2) when test data with imputed values are used in the selected model.

The expectation is the model works reliably if these two errors are close to each other. Although we know they cannot be equal in all implementations, we want to know how much they are close statistically (addressing RQ3). We use paired samples Wilcoxon test (a non-parametric alternative of paired t-test) to examine the equality of these two errors. The Wilcoxon test shows that these two errors are not equal statistically. However, they become equal if the values of the imputed test error reduce by 3%. So, the classification errors of the imputed test data are just 3% percent more than when there is no missing value in average. We use the difference of these errors to compare the performance of the model with different imputation objective functions. The average values of this difference are listed in TABLE IV.

TABLE IV. ERROR DIFFERENCE FOR OBJECTIVE FUNCTIONS

| Model with | ASW | Correlation | VR |
|---|---|---|---|
| Error difference | 0.05 | 0.049 | 0.05 |

TABLE IV shows that model's formulations with different imputation objective functions work similarly. We examine the impact of objective functions on error difference using Kruskal-Wallis test. The test verifies that the error differences of objective functions are statistically equal in average. We also examine the effect of missing combinations on error difference using Kruskal-Wallis test. The results show that the error difference is sensitive to missing ratios and patterns. The error difference increases as the ratio increase or pattern changes in order of simple, medium and complex.

*E. More investigations on objective functions*

We examine how well the objective functions can represent the imputation and prediction (addressing RQ1). To do this examination, we use metrics for imputation accuracy and prediction error. For each chromosome in all generations, we calculated imputation accuracy using MAE and RMSE and classification error of the test data when there is no missing value. Then, we investigate the relationship between objective functions and their relevant metrics using normalized mutual information (NMI). The results show there are high correlation between objective values and metrics (TABLE V).

TABLE V. NORMALIZED MUTUAL INFORMATION RESULTS

| Model | Objective function | NMI | | |
|---|---|---|---|---|
| | | MAE | RMSE | Test error |
| ASW | Imputation | 0.93 | 0.93 | - |
| | Model selection | - | - | 0.84 |
| Correlation | Imputation | 0.97 | 0.97 | - |
| | Model selection | - | - | 0.85 |
| VR | Imputation | 0.93 | 0.93 | - |
| | Model selection | - | - | 0.88 |

*F. Performance comparison*

In this section, we compare the performance of our model with [12] in terms of test classification error. In [12], researchers propose a multi-objective model to select the best subset for train data to improve the prediction accuracy of SVM classification models. The results are presented in TABLE VI.

We indicate the best value using underline. It should be noted that missing values exist in our model and the error values are the averaged values from missing combinations. Whereas, there is no missing value in [12]. The results show that our model lead to less and equal test error for zoo and sonar datasets respectively. In case of sonar dataset, the imputed test error is equal with the test error comes from the optimal TSS in [12]. These results validate our model's capability in model selection in presence of missing values. In addition, we can see the model performs consistently using different imputation objective functions.

TABLE VI. PERFORMANCE COMPARISON

| Data | Model ASW | | Model Correlation | | Model VR | | [12] |
|---|---|---|---|---|---|---|---|
| | Imputed test error | Test error | Imputed test error | Test error | Imputed test error | Test error | Test error |
| iris | 0.13 | 0.06 | 0.14 | 0.07 | 0.13 | 0.06 | 0.04 |
| zoo | 0.15 | 0.1 | 0.15 | 0.1 | 0.15 | 0.1 | 0.1 |
| sonar | 0.26 | 0.23 | 0.26 | 0.23 | 0.26 | 0.23 | 0.26 |

## V. CONCLUSION AND FURTHER RESEARCH

In this paper, we have proposed a multi-objective optimization model to do machine learning for incomplete data. The contributions of the paper to the machine learning area are listed below:

- The proposed model provides an opportunity to fill in the missing values at the same time of doing classification with no need of separate offline imputation.
- The proposed model selects the best model for classification automatically.
- Three objective functions are proposed to optimize the imputation process.
- The proposed model optimizes the parameters especially determines the best number of clusters internally.

We have designed extensive experiments under different scenarios to examine the performance of the proposed model and compare different formulations. These examinations and comparisons have been done via measuring imputation and classification metrics. The experimental results show that the proposed model is robust and perform well against different scenarios. We also provide insights on how missing ratios and patterns impact the performance of the proposed model using non-parametric tests. Furthermore, three formulations of imputation objective function show consistent and promising performance.

This study is an initial work on developing multi-objective optimization models for machine learning with incomplete data. Research directions can be suggested to improve the current study as follows. We have used fuzzy clustering and SVM method for imputation and classification respectively. In future, researchers can use different imputation and classification methods in optimization models and compare their performance. New objective functions for imputation and model selection can be proposed. The proposed model can be applied on medical data to examine its performance in solving medical problems. In addition, the multi-objective optimization model can be extended by introducing more objective functions to add more capability such as feather selection, TSS, etc.